\documentclass[10pt,twocolumn,letterpaper]{article}

\newif\ifarxiv \arxivtrue \ifarxiv \usepackage{cvpr} \else \usepackage[review]{cvpr} \fi        %
\definecolor{cvprblue}{rgb}{0.21,0.49,0.74}
\usepackage[
    pagebackref,
    breaklinks,
    colorlinks,
    allcolors=cvprblue
]{hyperref}

\def\confName{CVPR}
\def\confYear{2027}

\title{Reinforcement Learning for VLAs Requires Updating Only a Single Layer}
\author{
  Yueru Jia$^{1,}$\thanks{\hangindent=0.4em\hangafter=1\raggedright Equal contributions. \textsuperscript{\textdagger}Project lead. \texttt{perryd@stanford.edu},
  \texttt{jiayueru@stu.pku.edu.cn}}\quad
  Perry Dong\textsuperscript{2,\,*,\,\textdagger}\quad
  Dorsa Sadigh$^{2}$\quad
  Chelsea Finn$^{2}$\quad
  Shanghang Zhang$^{1}$\\[0.5em]
  \normalfont $^{1}$Peking University\quad $^{2}$Stanford University
}

\begin{document}\ifarxiv %
\maketitle

\begin{abstract}
Online reinforcement learning (RL) has shown promise
for improving vision-language-action (VLA) models
and unlocking new manipulation capabilities.
While selective VLA fine-tuning has been studied
under supervised learning, the trainable capacity
required during online RL remains less understood.
Unlike behavior cloning on fixed demonstrations,
the action distribution used to train the VLA
continually evolves through environment interaction.
This raises the question: How much of a pretrained
VLA must remain trainable during online RL?
To investigate this question, we conduct a systematic
layer-wise study in an online RL framework, where
the VLA is fine-tuned alongside an edit policy
and critic.
Across five LIBERO tasks, updating a single
action expert layer of $\pi_{0.5}$ at higher relative depth
achieves performance comparable to full VLA
transformer-block fine-tuning, with several
layers exhibiting competitive results.
Further analyses show that individual components
within a single layer can support online adaptation,
while localized updates induce attention
redistribution across downstream frozen layers,
revealing changes in network computation
beyond the updated layer.
We observe similar trends across two additional
VLA architectures, MolmoAct2 and GR00T-N1.6.
On three real-world precision manipulation tasks,
single-layer adaptation achieves 30/30 successful
trials per task, matching full VLA fine-tuning
under the same online RL framework.
\end{abstract}

\section{Introduction}
\label{sec:intro}

Vision-language-action (VLA) models have acquired strong
manipulation capabilities through large-scale
vision-language pretraining and diverse robotic
demonstrations~\cite{
kim2024openvlaopensourcevisionlanguageactionmodel,
intelligence2025pi05visionlanguageactionmodelopenworld,
brohan2023rt2visionlanguageactionmodelstransfer,
nvidia2025gr00tn1openfoundation,
li2024cogactfoundationalvisionlanguageactionmodel,
black2026pi0visionlanguageactionflowmodel}.
Online reinforcement learning (RL) offers a promising
approach to further improve these policies through
environment interaction and reward feedback~\cite{
lu2025vlarlmasterfulgeneralrobotic,
li2025simplevlarlscalingvlatraining,
chen2026pitextttrlonlinerlfinetuning,
dong2026expoftsampleefficientreinforcementlearning}. 
Although VLA fine-tuning has been extensively
studied under supervised learning~\cite{
kim2025finetuningvisionlanguageactionmodelsoptimizing,
kim2026adaptivecapacityallocationvision,
jiang2026vlagseboostingparameterefficientfinetuning,
nguyen2026finetuningvisionlanguageactionmodelsrequires,
syed2026layersneedtuningdiagnosing},
the trainable capacity required for effective
online VLA adaptation with reinforcement learning remains less understood.
In this paper, we ask: \emph{How much of a pretrained
VLA must remain trainable for effective online
adaptation through reinforcement learning?}

\begin{figure*}[t]
    \centering
    \includegraphics[width=\textwidth]
        {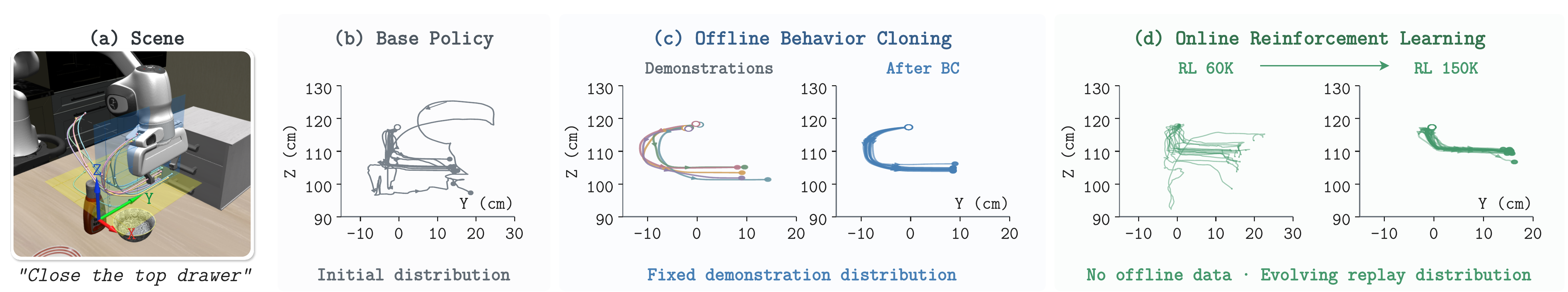}
    \caption{
    \textbf{Illustration of the different adaptation dynamics in
    offline BC and online RL.} Starting from a pretrained
    policy (b) in the scene shown in (a), offline BC fits a
    fixed demonstration distribution (c). In online RL,
    the data distribution evolves as updated policies collect
    new experience through online sampling (d).
    }
    \label{fig:task28_single_layer_behavior}
\end{figure*}

Previous studies have explored layer pruning and
selective fine-tuning for efficient VLA adaptation
under behavior cloning~\cite{
nguyen2026finetuningvisionlanguageactionmodelsrequires,
syed2026layersneedtuningdiagnosing}.
Unlike behavior cloning on fixed demonstrations,
a policy's action distribution continually evolves
through interaction during online RL.
Fig.~\ref{fig:task28_single_layer_behavior}
illustrates this evolving learning process through
a representative drawer-closing task.
While BC learns from a fixed set of demonstration
trajectories, the online RL policy exhibits
substantially different execution trajectories
as training progresses.
Crucially, changes in policy behavior also influence
subsequent experience collection, continually
reshaping the action supervision used for future
VLA updates.
We investigate this question under the EXPO-FT
online RL framework~\cite{
dong2026expoftsampleefficientreinforcementlearning},
where the VLA learns from replayed actions through
flow matching, while an edit policy and critic
guide action generation and shape subsequent
experience collection.
Since policy improvement emerges from this joint
learning process, it remains unclear whether
extensive VLA updates are necessary or whether
a small trainable subset can provide sufficient
adaptation capacity.

To investigate this question, we conduct a systematic
layer-wise study of $\pi_{0.5}$ within the EXPO-FT
framework. Modern VLAs often couple multimodal representations
from a vision-language backbone with action generation
performed by an action expert.
We examine both the vision-language backbone and
the action expert by restricting trainable VLA
parameters to one original transformer block at
a time, while keeping the remaining weights frozen.
This controlled design preserves the complete
model architecture and allows us to examine
single-layer adaptation within the
online RL learning process, where changes in
policy behavior continually reshape subsequent
training experience. It also enables us to investigate
whether a single layer provides sufficient
adaptation capacity for improvement with online RL.

We evaluate this setting across five LIBERO
tasks~\cite{liu2023benchmarkingknowledgetransfer} for online RL.
Our layer-wise results show that several
action expert layers at higher relative depths
achieve performance comparable to full
fine-tuning, while the Gemma backbone exhibits greater
task-dependent sensitivity to layer choice.
These findings suggest that a single layer
can provide sufficient capacity for online
adaptation without updating the entire
pretrained VLA.
We evaluate single-layer adaptation on
two more representative VLA models,
MolmoAct2~\cite{
fang2026molmoact2actionreasoningmodels}
and GR00T-N1.6~\cite{
nvidia2025gr00tn16},
observing similar effectiveness across architectures. Having observed that a single transformer layer
can support online adaptation, we further
investigate whether updates can be restricted
to individual modules and how localized changes
affect computation across the network.
Module ablations show that attention, feed-forward
network (FFN), and AdaRMS updates can independently
support effective adaptation, indicating that
updating an entire transformer block is not always
necessary.
Meanwhile, attention analysis reveals changes in
attention allocation across downstream frozen layers.
Together, these findings highlight that localized
updates can support adaptation while influencing
computation beyond the updated layer.

To evaluate single-layer adaptation in real-world
settings, we conduct three precision manipulation
tasks using a Franka Research 3 robot, covering
precise alignment, articulated-object manipulation,
and contact-rich tool use.
These tasks examine whether localized adaptation
remains effective during precise
object interactions and contact.
Single-layer adaptation achieves 30/30 successful
trials on each task, matching full fine-tuning
under EXPO-FT while outperforming other state-of-the-art real-world
RL approaches.
These results demonstrate the effectiveness of
localized VLA adaptation in physical manipulation.

Our contributions are summarized as follows.
We conduct a systematic layer-wise study of how much
VLA adaptation is needed for effective online RL.
Across five LIBERO tasks, we find that a single
action expert layer at higher relative depth
can approach full fine-tuning performance.
We further investigate the sufficiency of
individual modules and analyze attention
redistribution across frozen layers.
Finally, we extend our findings to two additional
VLA architectures and validate single-layer
adaptation on three real-world precision
manipulation tasks.
\section{Related Work}

\subsection{Reinforcement Learning With VLAs}

Reinforcement learning (RL) has been widely studied
for robotic manipulation, with recent real-world
systems demonstrating the effectiveness of experience
reuse and human feedback
~\cite{luo2025precisedexterousroboticmanipulation,
luo2025serlsoftwaresuitesampleefficient,
lei2026rl100performantroboticmanipulation,
luo2024rlifinteractiveimitationlearning}.
Applying RL to pretrained VLAs introduces additional
challenges due to their expressive policy backbone.
Existing works explore direct VLA fine-tuning through
RL objectives tailored to the VLA or diffusion and flow-based
policies~\cite{dong2025expostable,dong2026expoftsampleefficientreinforcementlearning,dong2026reinforcementlearningrealtimevisionlanguageaction,intelligence2025pi06vlalearnsexperience,xu2026rltokenbootstrappingonline,wagenmaker2025steeringdiffusionpolicylatent,chen2026pitextttrlonlinerlfinetuning,lu2025vlarlmasterfulgeneralrobotic,li2025simplevlarlscalingvlatraining}.
Several approaches improve pretrained robot policies
by learning auxiliary controllers or value functions
while keeping the base policy frozen during subsequent
policy improvement
~\cite{wagenmaker2025steeringdiffusionpolicylatent,xu2026rltokenbootstrappingonline}.
EXPO-FT
~\cite{dong2026expoftsampleefficientreinforcementlearning}
combines improvement from value maximization with a separate edit policy with stable supervised updates to the base VLA, enabling stable online learning.
Building on EXPO-FT, we investigate how much of the VLA
needs to be updated within this jointly learned system,
showing that adaptation through a single original
transformer block can achieve performance comparable
to full VLA fine-tuning.

\begin{figure*}[t]
    \centering
    \includegraphics[width=\textwidth]
    {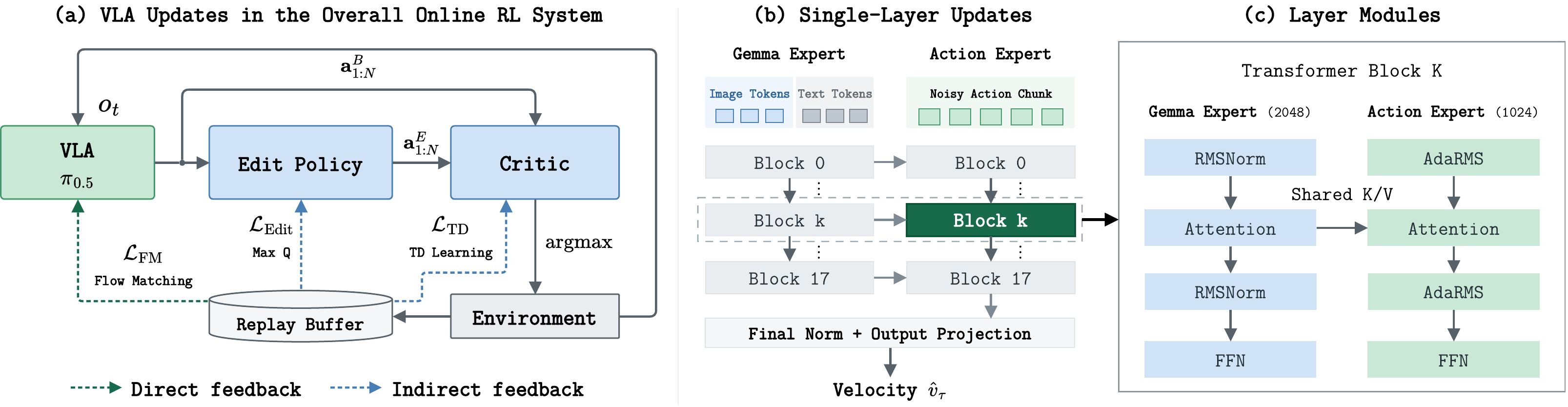}
    \caption{
\textbf{Overview of single-layer VLA adaptation under EXPO-FT.}
(a) Online RL with flow-matching VLA updates.
(b) One trainable transformer block in the vision-language
backbone or action expert, with full pretrained computation retained.
(c) Illustration of the selected block's internal modules,
highlighting attention.
    }
    \label{fig:method_overview}
\end{figure*}

\subsection{Efficient VLA Adaptation}

Efficient VLA adaptation under behavior cloning (BC) has been studied from both training and action-sampling perspectives. On the training side, methods reduce optimization costs through low-rank fine-tuning, structured parameter updates, adaptive capacity allocation, and selective component training~\cite{hu2021loralowrankadaptationlarge,kim2024openvlaopensourcevisionlanguageactionmodel,kim2026adaptivecapacityallocationvision,jiang2026vlagseboostingparameterefficientfinetuning,driess2026knowledge,ferchau2026efficiencylorafinetuningvisionlanguageaction,R_mer_2026}. On the sampling side, approaches improve action generation efficiency through optimized action representations and decoding strategies~\cite{kim2025finetuningvisionlanguageactionmodelsoptimizing}. {For online RL, FASTER reduces sampling costs by learning to filter action candidates during denoising~\cite{dong2026fastervalueguidedsamplingfast}.} Another line of work investigates layer-wise redundancy and selective adaptation through layer pruning and region-specific tuning~\cite{nguyen2026finetuningvisionlanguageactionmodelsrequires,syed2026layersneedtuningdiagnosing,liu2026layermysteryvlainformationtheoretical}. However, trainable capacity in online VLA RL remains less explored. The most closely related work to ours is Low-Rank VLA RL~\cite{oh2026low}, which analyzes the structure and functional contribution of PPO-induced parameter updates, particularly in timestep-conditioning modules. In contrast, our work investigates the trainable capacity of individual transformer layers under a more sample efficient off-policy online RL framework, examining whether restricting updates to a single original layer from the start of online training can support effective policy improvement.
\section{Single-Layer Adaptation Capacity}
\label{sec:single_layer_study}

In this section, we investigate the trainable capacity required for online VLA adaptation through a systematic layer-wise study under an online RL framework, as shown in Fig.~\ref{fig:method_overview}. Each run updates one original transformer block in the vision-language backbone or action expert while retaining the complete pretrained computation. We compare performance with full fine-tuning and evaluate cross-backbone applicability.

\subsection{Preliminaries}
\label{sec:preliminaries}

\noindent\textbf{Flow-Based VLA Policy.}
We adopt $\pi_{0.5}$~\cite{intelligence2025pi05visionlanguageactionmodelopenworld},
a mixture-of-transformers (MoT) comprising a PaliGemma backbone
and a flow-matching action expert
(Fig.~\ref{fig:method_overview}(b)).
The Gemma language backbone and the action expert each have 18 transformer layers, with hidden
dimensions of 2048 and 1024, respectively.
Gemma processes the visual, language, and state prefix,
while the action expert processes noisy action tokens.
Joint masked self-attention allows action tokens to attend
to both the prefix and other action tokens, with information
flowing from Gemma to the action expert. Given an action chunk $\mathbf{a}$, noise
$\boldsymbol{\epsilon}\sim\mathcal{N}(\mathbf{0},\mathbf{I})$,
and timestep $\tau$, we form
$\mathbf{x}_{\tau}=(1-\tau)\mathbf{a}+\tau\boldsymbol{\epsilon}$
and train the velocity field with

\begin{equation}
    \mathcal{L}_{\mathrm{FM}}(\theta)
    =
    \mathbb{E}_{\mathbf{o},\mathbf{a},
    \tau,\boldsymbol{\epsilon}}
    \left[
    \left\|
    v_{\theta}(\mathbf{x}_{\tau},\tau,\mathbf{o})
    -
    (\boldsymbol{\epsilon}-\mathbf{a})
    \right\|_2^2
    \right],
    \label{eq:flow_matching}
\end{equation}

where $\mathbf{o}$ denotes visual, language, and state
observations. Action sampling integrates the velocity
field from noise at $\tau=1$ to actions at $\tau=0$.
A shared Time MLP and layer-specific adaptive RMS
normalization (AdaRMS) condition the action expert on $\tau$.

\noindent\textbf{EXPO-FT}~\cite{dong2026expoftsampleefficientreinforcementlearning}.
We build upon EXPO-FT,
an online off-policy RL framework combining a pretrained
VLA, a critic, and an edit policy that maximizes Q-value
(Fig.~\ref{fig:method_overview}(a)).
At each sampling step, the VLA proposes action chunks,
the edit policy transforms the actions toward higher Q-values, and the
critic evaluates original and edited candidates to select
an action for execution. The resulting self-collected
experience is stored in a replay buffer for subsequent training.
The critic learns through temporal-difference updates,
and the edit policy learns to maximize the Q-values. The base VLA learns from replayed actions
through the flow-matching objective in
Eq.~\eqref{eq:flow_matching}.
The VLA thus fits an evolving action distribution induced
by the system's reward-guided interactions.
We update the edit policy, critic, and critic image encoder following the EXPO-FT protocol.

\begin{figure*}[t]
    \centering
    \includegraphics[width=\textwidth]
{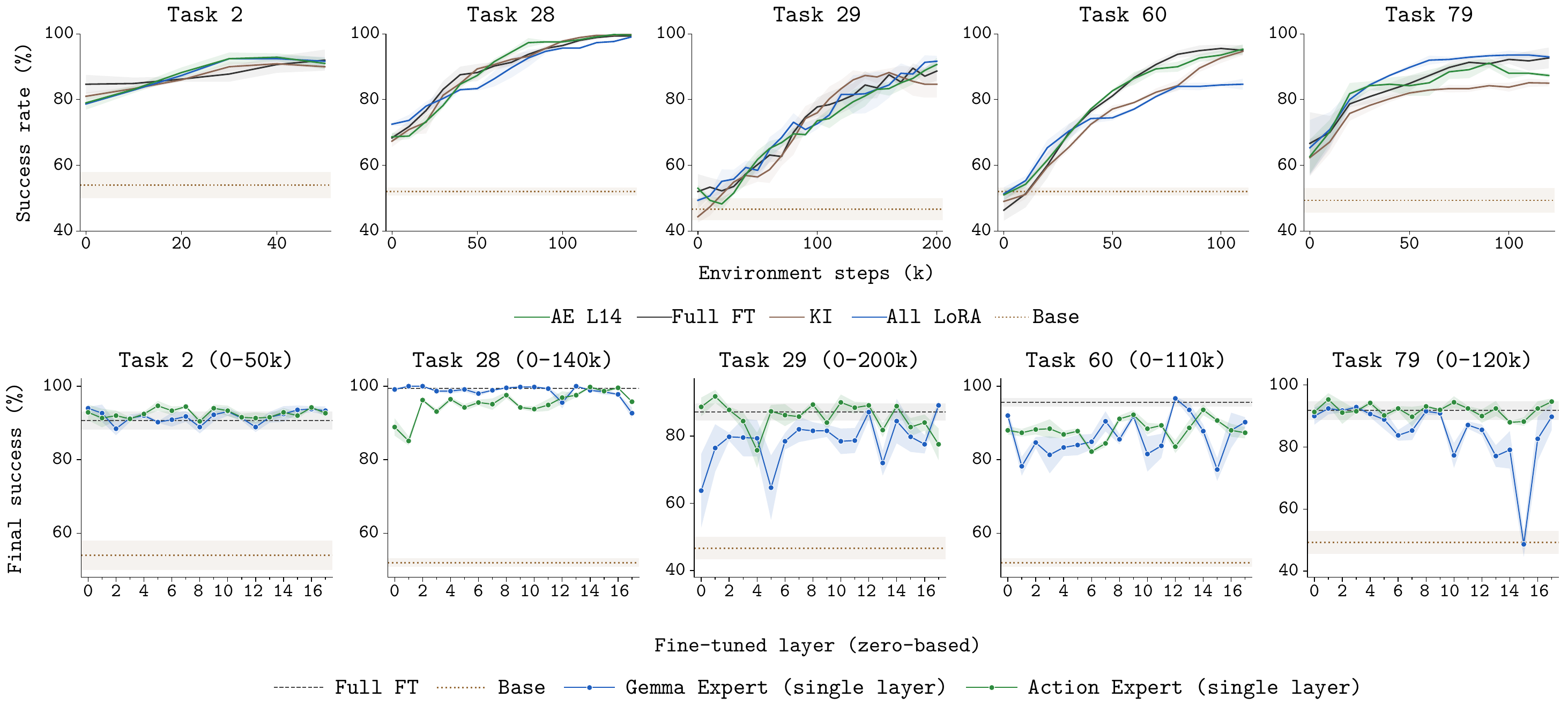}
\caption{
    \textbf{Single-layer VLA adaptation across five LIBERO tasks.}
    The top row compares AE L14 with adaptation baselines over environment steps.
    The bottom row reports final success rates across individual Gemma and action expert
    layers. Base denotes the initial VLA without online adaptation.
    Shaded regions indicate one standard error across three seeds.
}

    \label{fig:baseline_learning_curves}
    \label{fig:layerwise_results}
    \label{fig:layerwise_final_success}
\end{figure*}

\subsection{Adaptation Protocol}
\label{sec:single_layer}

We independently adapt blocks L0--L17 in each stack.
Let $\theta_l^s$ denote block $l$ in Gemma ($s=G$)
or the action expert ($s=A$). Each run uses
\begin{equation}
    \Theta_{\mathrm{train}}^{(s,l)}
    =
    \{\theta_l^s\},
    \qquad
    \Theta_{\mathrm{frozen}}^{(s,l)}
    =
    \Theta_{\mathrm{VLA}}
    \setminus\{\theta_l^s\}.
    \label{eq:single_layer}
\end{equation}
The trainable layer index remains fixed throughout each
run, and its original weights receive updates using
Eq.~\eqref{eq:flow_matching}.
All remaining VLA parameters are frozen, including
the other transformer blocks, visual encoder, token
embeddings, final normalization, action projections,
and shared Time MLP. Auxiliary RL components follow
the standard EXPO-FT updates.
LIBERO training uses autonomous environment interaction
without human action corrections. Real-robot training uses
the human-in-the-loop setup in Sec.~\ref{sec:real_world}.
The implementation separates the transformer stack into
a frozen prefix, the selected block, and a frozen suffix.
Prefix outputs are detached from automatic differentiation,
while input gradients propagate through the suffix to
update the selected block. Gradients and optimizer states
are allocated for this block.
We measure update time and memory to quantify training cost.

\subsection{Layer-Wise Evaluation}
\label{sec:layerwise_results} 
\label{sec:exp_setup}

\noindent\textbf{Experimental Setup.}
We initialize $\pi_{0.5}$ from OpenPI's
\texttt{pi05\_libero} checkpoint, fine-tuned on
the Spatial, Object, Goal, and Long suites of
LIBERO~\cite{liu2023benchmarkingknowledgetransfer}.
Following prior work~\cite{dong2026fastervalueguidedsamplingfast},
we select five LIBERO-90 tasks (2, 28, 29, 60, and 79) with initial
success rates of 40--60\%.
All online RL configurations start from the same
checkpoint with \textbf{zero task demonstrations}. We use an action chunk
length of $H=10$ and three seeds per configuration.
Additional training configurations and
implementation details are provided in the Appendix.

\noindent\textbf{Reference Configurations.}
We compare single-layer adaptation with
mainstream VLA fine-tuning strategies under
online RL, including the following configurations.

\begin{itemize}

    \item \textbf{Base:}
    Evaluates the initial SFT policy as a standalone
    VLA without online RL adaptation.

    \item \textbf{Full FT:}
    Jointly updates all transformer blocks in
    Gemma and the action expert during online RL.

    \item \textbf{Knowledge Insulation (KI)}~\cite{driess2026knowledge}:
    Freezes the vision-language backbone and
    fine-tunes the entire action expert.

    \item \textbf{All LoRA}~\cite{hu2021loralowrankadaptationlarge}:
    Trains low-rank adapters across both the
    vision-language backbone and action expert
    while freezing the pretrained weights.

\end{itemize}

For LIBERO, we evaluate each checkpoint over 50 episodes per seed. We use final success rate as the primary metric, defined
as the mean over the last $K=3$ recorded evaluation
checkpoints within each task's interaction budget,
then averaged over three seeds.
\begin{equation}
    F = \frac{1}{9}\sum_{r=1}^{3}\sum_{j=0}^{2}
    S_r\!\left(u_{r,N_r-j}\right).
    \label{eq:final_success}
\end{equation}
where $S_r(u)$ denotes the evaluation success rate (in percent), and
$u_{r,N_r}$ denotes the final recorded evaluation checkpoint
within the interaction budget. Final success is computed using
the unsmoothed evaluation scores. Shaded regions indicate one
standard error across three seeds. Learning curves show the
mean across three seeds after centered three-point smoothing.

\begin{figure*}[t]
    \centering
    \includegraphics[width=\textwidth]
        {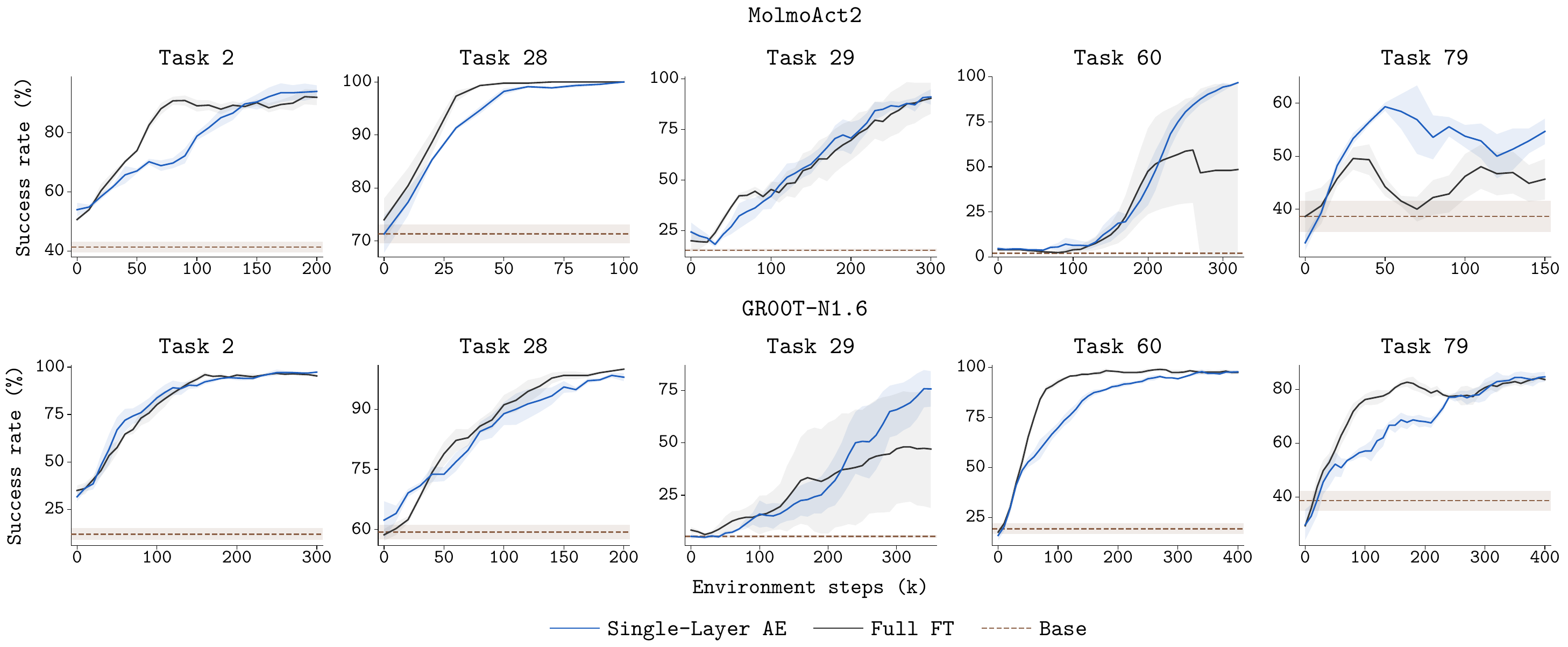}
\caption{
    \textbf{Learning curves of MolmoAct2 and GR00T-N1.6
    under EXPO-FT across five LIBERO tasks.}
    The top and bottom rows correspond to MolmoAct2
    and GR00T-N1.6, respectively. Each compares single-layer
    action expert adaptation with Full FT and the
    initial base policy over environment steps.
    Curves show mean performance across three seeds
    with centered three-point smoothing, and shaded regions
    indicate pointwise standard errors across three seeds.
}
    \label{fig:cross_backbone_learning_curves}
\end{figure*}

\noindent\textbf{Results.}
The top row of Fig.~\ref{fig:baseline_learning_curves}
compares AE L14 with Full FT, KI, and All LoRA
over environment steps.
AE L14 achieves final success comparable to
Full FT across five tasks, outperforming
it on Tasks 2, 28, and 29.
The bottom row presents the layer-wise results,
with learning curves provided in the Appendix.
Several action expert layers, particularly
L13--L16, achieve competitive performance,
indicating that the results are not limited
to L14.

Moreover, action expert adaptation exhibits
relatively stable performance across layers,
whereas Gemma is more sensitive to layer
selection, particularly on Tasks 29 and 79.
On Task 79, final success ranges from
88.00\% to 95.33\% for the action expert,
compared with 48.67\% to 92.89\% for Gemma.
Together, these results suggest that
a single action expert layer can provide
sufficient capacity for online adaptation,
with consistent performance across several
different update locations.

%
\subsection{Generalization Across VLA Backbones}
\label{sec:cross_model}

To examine whether the effectiveness of single-layer
adaptation extends beyond $\pi_{0.5}$, we evaluate two
additional flow-based VLAs, MolmoAct2 and GR00T-N1.6,
on the same five LIBERO tasks.
Both combine a vision-language backbone with an action
diffusion transformer (DiT), but differ in network depth
and conditioning mechanisms.
For each model, we select a single action expert layer
from a higher-depth region, using the same initial
checkpoint as the corresponding Full FT baseline
and no task demonstrations during online RL.
Detailed adaptation configurations are provided
in the Appendix.

\begin{itemize}
    \item \textbf{MolmoAct2}~\cite{
    fang2026molmoact2actionreasoningmodels}
    combines a 36-layer Molmo2-ER language backbone
    with a 36-layer action expert.
    Each action block performs self-attention over
    action representations and cross-attention to
    the projected K/V cache of the corresponding
    VLM layer, providing explicit layer-wise
    cross-modal conditioning. We fine-tune action expert block L27, whose hidden size is 768.

    \item \textbf{GR00T-N1.6}~\cite{
    nvidia2025gr00tn16}
    combines a Cosmos-2B-based vision-language
    backbone with a 32-block action DiT.
    Its action blocks alternate self-attention
    over state and action representations with
    cross-attention to shared VLM features,
    providing shared conditioning across depth. We fine-tune block L26, a cross-attention block with hidden size 1,536.
\end{itemize}

\noindent\textbf{Results.}
Fig.~\ref{fig:cross_backbone_learning_curves}
shows that single-layer action expert adaptation
achieves competitive performance across both
VLA backbones.
On MolmoAct2, it remains competitive with Full FT
on all five tasks at the end of training.
In particular, on Task 60, Full FT initially
improves faster but plateaus at a lower success
rate, whereas single-layer adaptation continues
improving and eventually surpasses it.
A similar advantage appears on Task 79, where
single-layer adaptation maintains higher success
throughout much of training.
On GR00T-N1.6, it achieves comparable final
performance on Tasks 2, 28, 60, and 79,
while substantially outperforming Full FT
on Task 29.
These differences indicate that updating more
layers does not necessarily yield faster
convergence or better final performance.
\begin{figure*}[t]
\centering
\includegraphics[width=\textwidth]{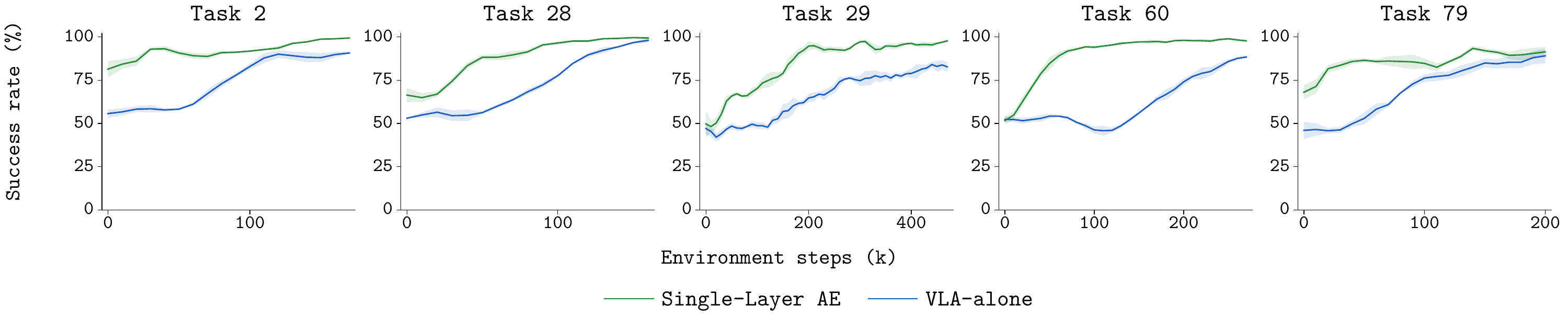}
\caption{
    \textbf{Policy improvement within the VLA during online adaptation.}
    Success rates of single-layer L14 EXPO-FT and VLA-alone
    across five LIBERO tasks. Curves show mean performance
    across three seeds with centered three-point smoothing;
    shaded regions indicate pointwise standard errors
    across three seeds.
}
\label{fig:policy_improvement_vla}
\end{figure*}

\section{Understanding Localized Adaptation With Single-Layer Updates}
\label{sec:adaptation_analysis}

Our layer-wise results show that updating a single
action expert layer achieves performance comparable
to Full FT while modifying only a small fraction
of VLA parameters. To understand why localized
adaptation remains effective, we examine three
questions. First, we assess whether online RL
improves the VLA itself in
Sec.~\ref{sec:policy_improvement_vla}.
Next, we analyze how single-layer updates
redistribute attention across the network,
including downstream frozen layers, in
Sec.~\ref{sec:attention_analysis}.
Finally, we examine which components of the
adapted layer support effective adaptation
in Sec.~\ref{sec:module_ablation}.

\subsection{Policy Improvement within the VLA}
\label{sec:policy_improvement_vla}

During online RL, the single-layer VLA policy
is jointly optimized with an edit policy
and the critic. Improvements in the
complete system do not necessarily indicate
that the VLA itself has acquired stronger action
generation capabilities. To investigate whether
these improvements are reflected in the VLA
and support standalone deployment, we evaluate
the single-layer policy in two execution modes:
VLA-alone, which directly executes VLA-predicted
actions, and the complete single-layer system,
which employs the learned edit policy and critic.
For this analysis, we use a separate set of training runs under the same experimental settings as the layer-wise study.

As shown in Fig.~\ref{fig:policy_improvement_vla},
VLA-alone progressively improves across all five
LIBERO tasks. On Tasks 2 and 28, its success
rates gradually approach those of the complete
single-layer system, despite slower initial
improvement. On Task 79, both execution modes
achieve comparable performance toward the end
of training. Tasks 29 and 60 also exhibit
substantial improvements in standalone VLA
performance. These results suggest that
single-layer adaptation effectively enhances
the VLA's own action generation capabilities,
enabling standalone deployment.

\subsection{Attention Redistribution}
\label{sec:attention_analysis}

To understand how localized updates influence
the frozen network, we analyze changes in attention
allocation across transformer layers.
Specifically, we measure the attention mass from
action queries to four token groups: base-camera
images, wrist-camera images, language, and actions.
For token group $g$, measured layer $m$, and
adaptation configuration $c$, we define
\begin{equation}
\begin{aligned}
M_g^{(c,m)}
&=
\frac{1}{|\mathcal{T}|H_m|\mathcal{Q}|}
\sum_{\tau\in\mathcal{T}}
\sum_{h=1}^{H_m}
\sum_{q\in\mathcal{Q}}
\sum_{k\in\mathcal{K}_g}
A_{qk}^{(c,m,h,\tau)},\\
\Delta M_g^{(c,m)}
&=
100\left(
M_g^{(c,m)}
-
M_g^{(\mathrm{base},m)}
\right),
\end{aligned}
\label{eq:attention_mass}
\end{equation}

where $A_{qk}^{(c,m,h,\tau)}$ denotes the masked
post-softmax attention weight, $\mathcal{Q}$
and $\mathcal{K}_g$ denote valid action queries
and keys belonging to token group $g$, respectively,
and $H_m$ is the number of attention heads.
We evaluate three flow timesteps,
$\mathcal{T}=\{0.90,0.50,0.10\}$, and report
the resulting attention-mass changes in percentage
points relative to the pretrained VLA.

\begin{figure}[t]
    \centering
    \includegraphics[width=\linewidth]
        {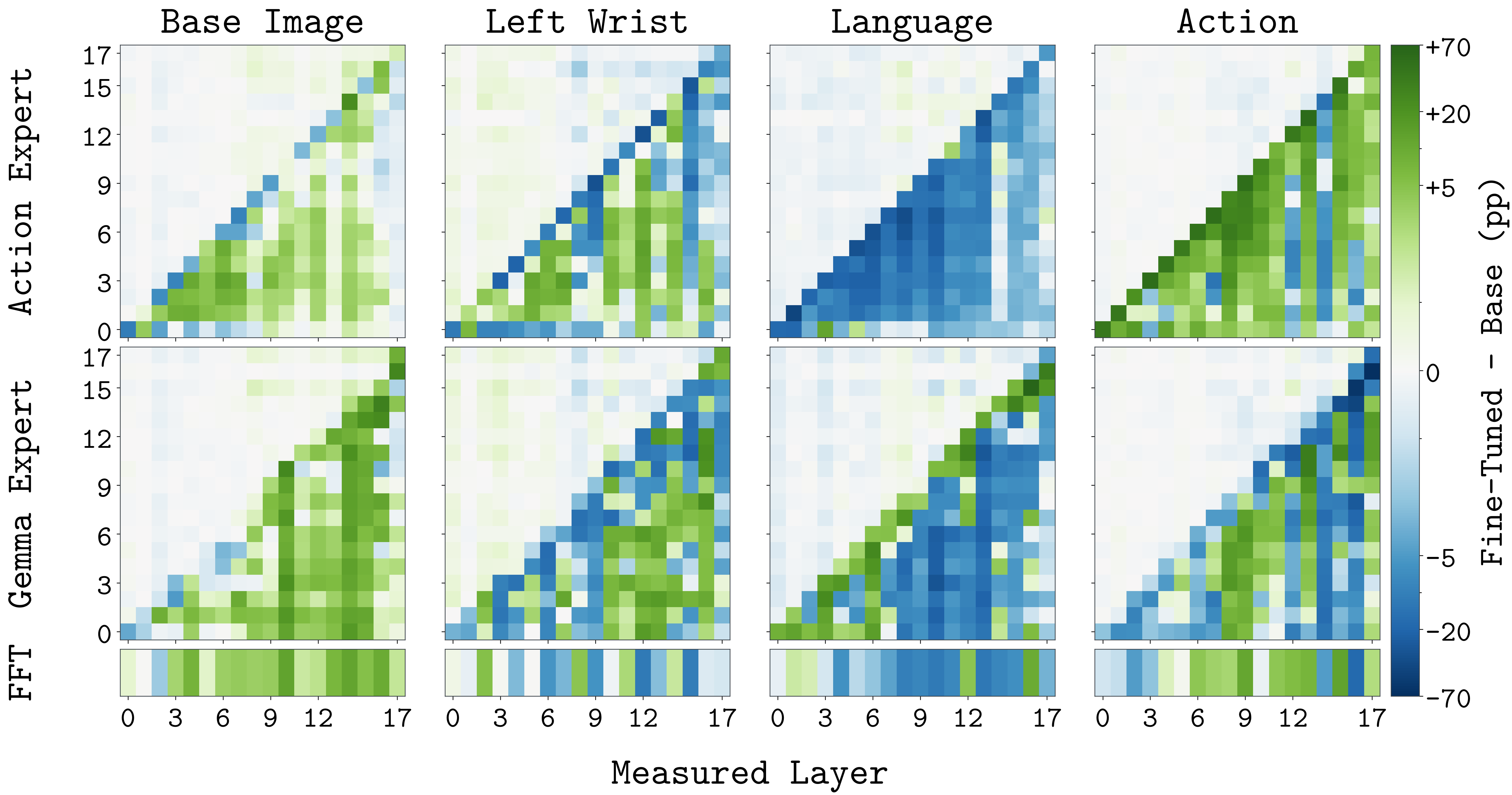}
\caption{
    \textbf{Attention redistribution under single-layer and full fine-tuning.}
    Attention-mass changes relative to the pretrained VLA
    across single-layer Gemma, single-layer action expert,
    and Full FT on LIBERO Task 28.
}
    \label{fig:attention_mass}
\end{figure}

Fig.~\ref{fig:attention_mass} visualizes the
attention changes induced by adapting each of
the 36 transformer layers individually
(18 Gemma and 18 action expert layers) on
LIBERO Task 28. The horizontal and vertical
axes indicate measured and adapted layers,
respectively, while green and blue represent
increases and decreases in attention mass.

Single-layer action expert adaptation
produces attention redistribution patterns
that qualitatively resemble those of Full FT,
with increased attention to action tokens
and reduced attention to language tokens.
These changes propagate to
subsequent layers whose parameters remain
frozen. In contrast, Gemma expert adaptation
exhibits more heterogeneous patterns across
token groups and network depth.
These observations suggest that updating
a single action expert layer can reshape
attention allocation throughout the network,
even without modifying the downstream layers.

\subsection{Module Ablations}
\label{sec:module_ablation}

Motivated by the attention redistribution
observed above, we further investigate which
components within a single action expert layer
contribute to effective online adaptation.
Using the selected action expert layer L14,
we compare whole-block, attention-only,
FFN-only, and AdaRMS-only updates, along with
ablations of individual attention projections.
All variants share the same pretrained
initialization and EXPO-FT hyperparameters.
Fig.~\ref{fig:module_ablation} reports
mean performance across three seeds with centered
three-point smoothing and pointwise standard errors.
Final success rates are computed
following Eq.~\eqref{eq:final_success}.

\begin{figure}[t]
    \centering
    \includegraphics[width=\linewidth]
        {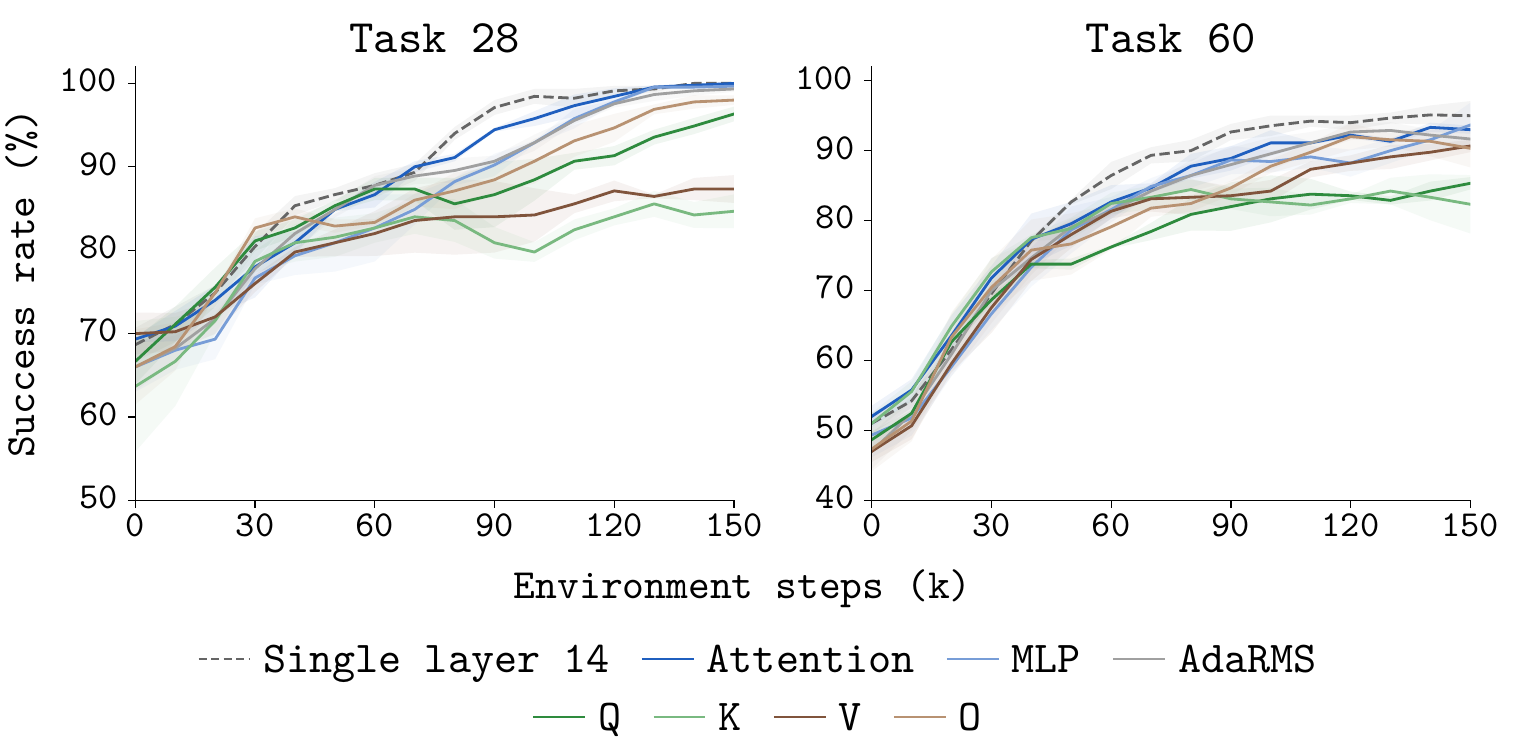}
\caption{
    \textbf{Module ablations within action expert
    layer L14 on LIBERO tasks 28 (left) and 60 (right).}
    Each setting uses three seeds, with shading indicating one standard error.
}
    \label{fig:module_ablation}
\end{figure}

As shown in Fig.~\ref{fig:module_ablation},
attention-only adaptation achieves performance
close to whole-block adaptation across both tasks,
despite updating fewer parameters.
On Task 28, all three module-only variants
achieve final success rates of 99.11--99.78\%,
while on Task 60, their results remain within
3.56 percentage points of whole-block adaptation.
Among individual attention projections,
the output projection performs consistently
well across both tasks, whereas other projections
show task-dependent effectiveness.

The strong performance of FFN-only and
AdaRMS-only adaptation further indicates
that effective updates are not restricted
to the attention module.
Together with the attention redistribution
analysis, these findings suggest that a single
action expert layer provides multiple effective
adaptation pathways, with attention offering
a particularly effective mechanism for
modifying action generation.

\begin{figure*}[t]
    \centering
        \includegraphics[width=\textwidth]
        {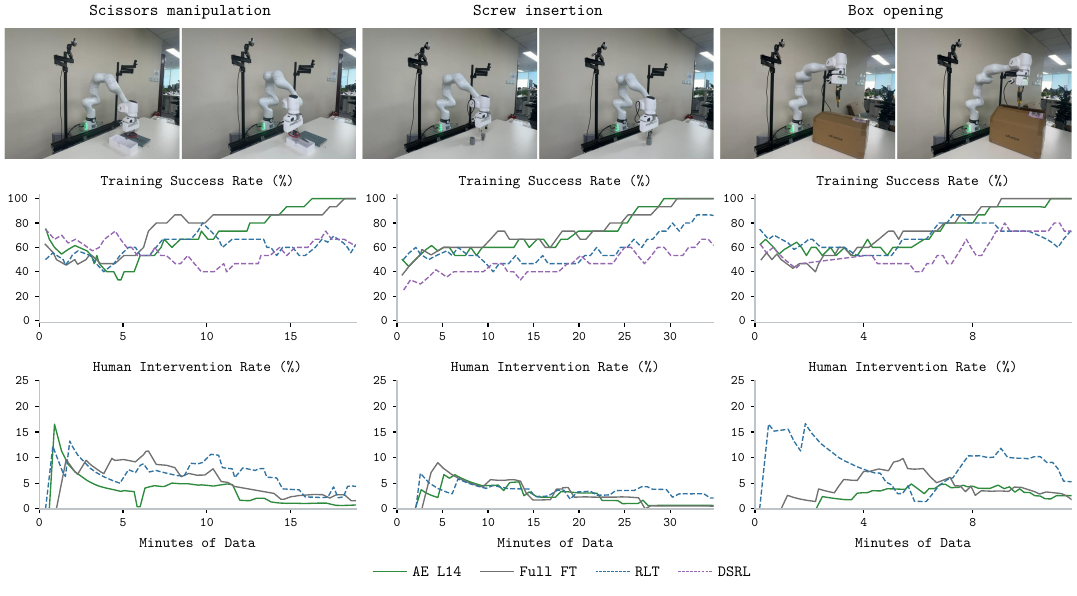}
\caption{
\textbf{Three real-robot tasks and their online training dynamics.} Columns show \textit{scissors manipulation},
\textit{screw insertion}, and \textit{box opening}.
The top row presents task snapshots. The middle row compares training success across
AE L14, Full FT, RL Token, and DSRL, while the
bottom row shows human intervention rates for
AE L14, Full FT, and RL Token.
}
 
\label{fig:real_robot_tasks}
\end{figure*}

\section{How Does Single-Layer Adaptation Work on Real-World Robots?}
\label{sec:real_world}

We compare single-layer adaptation with Full FT and other
baselines on three real-world manipulation tasks, using
the representative late action expert layer selected for
its strong simulation performance in
Sec.~\ref{sec:layerwise_results}.

\noindent\textbf{Experimental Setup.}
A Franka Research 3 arm receives images from an Intel
RealSense D405 wrist camera and D435 third-person camera,
together with proprioceptive observations.
A local NVIDIA RTX 5090 workstation handles control
and data collection, while a remote NVIDIA A800 server
performs inference and optimization.
Real-robot training uses offline task demonstrations and
human-in-the-loop online data collection under EXPO-FT,
updating the selected action expert layer and the auxiliary
RL components while freezing the remaining VLA parameters.

During online training, a human operator may take control
to provide corrective actions when necessary.
The executed corrections are stored in the replay buffer
and used for subsequent learning.
Single-layer adaptation and Full FT follow the same
human-in-the-loop EXPO-FT protocol.
Final policies are evaluated autonomously without human
action corrections.
Additional experimental details are provided
in the Appendix.

\noindent\textbf{Task Description.}
We design three real-world precision-manipulation
tasks spanning insertion, articulated-object
manipulation, and tool use
(Fig.~\ref{fig:real_robot_tasks}).
For each task, we collect 10--20 offline demonstrations
to fine-tune the pretrained VLA, obtaining a task-specific
SFT policy that serves as the initialization for
subsequent online RL adaptation.

\begin{itemize}
    \item \textit{Scissors manipulation.}
    The robot closes the scissors, transports them to a
    box, and places them inside. The initially open
    scissors require precise gripper pose adjustment
    and closing-height control for a stable grasp.

    \item \textit{Screw insertion.}
    The robot picks up a screw, aligns it with a nut,
    and inserts it into the opening. The screw's
    polygonal geometry constrains the grasp pose,
    while tightly coupled position and orientation
    control is needed to avoid collisions during insertion.

    \item \textit{Box opening.}
    The robot uses a utility knife to cut open a
    cardboard box. The blade must accurately follow
    the narrow taped seam, as slight deviations can
    damage the cardboard instead of opening the box.
\end{itemize}

\begin{table}[t]
\centering
\caption{
\textbf{Real-world manipulation success rates across three tasks.}
Results are reported as successful trials out of 30,
with average success rates (\%) in the last column.
}
\label{tab:real_world_success}
\setlength{\tabcolsep}{9pt}
\renewcommand{\arraystretch}{1.2}
\resizebox{\columnwidth}{!}{
\begin{tabular}{@{}l|ccc|c@{}}
\hline
Method & Scissors & Screw & Box & Avg. (\%) \\
\hline
SFT~\cite{intelligence2025pi05visionlanguageactionmodelopenworld}
& 14/30 & 21/30 & 13/30 & 53.33 \\

DSRL~\cite{wagenmaker2025steeringdiffusionpolicylatent}
& 23/30 & 25/30 & 22/30 & 77.78 \\

RL Token~\cite{xu2026rltokenbootstrappingonline}
& 21/30 & 29/30 & 24/30 & 82.22 \\

Full FT~\cite{dong2026expoftsampleefficientreinforcementlearning}
& \textbf{30/30}
& \textbf{30/30}
& \textbf{30/30}
& \textbf{100.00} \\
\hline
Single AE L14
& \textbf{30/30}
& \textbf{30/30}
& \textbf{30/30}
& \textbf{100.00} \\
\hline
\end{tabular}
}
\end{table}

\noindent\textbf{Baselines.}
We compare single-layer adaptation with the initial
SFT policy and three online RL baselines, all using
$\pi_{0.5}$ as the base VLA:
\begin{itemize}
\interlinepenalty=10000

    \item \textbf{SFT:}
    The pretrained VLA is fine-tuned on task-specific
    demonstrations and evaluated without online RL.

    \item \textbf{DSRL:}~\cite{wagenmaker2025steeringdiffusionpolicylatent}
    The pretrained VLA remains frozen during online RL.
    A lightweight MLP-based actor-critic is trained
    through off-policy RL, where the actor selects
    observation-conditioned initial noise and the critic
    provides value estimates to guide its optimization.
    This learned noise policy steers action generation
    without updating the VLA weights.

    \item \textbf{RL Token:}~\cite{xu2026rltokenbootstrappingonline}
    A compact RL-token representation is learned by
    reconstructing pretrained VLA features.
    During online RL, the VLA and representation module
    remain frozen, while a lightweight MLP actor-critic
    learns to refine VLA action proposals through
    value optimization and action regularization.

    \item \textbf{Full FT (EXPO-FT):}~\cite{dong2026expoftsampleefficientreinforcementlearning}
    All transformer blocks in the vision-language
    backbone and action expert are updated, with
    the MLP-based edit policy and critic trained jointly.

\end{itemize}

\noindent\textbf{Results.}
As shown in Tab.~\ref{tab:real_world_success},
single-layer AE L14 achieves 30/30 successes on
all three real-world tasks, matching Full FT
and improving average success from 53.33\%
for the initial SFT policy to 100.00\%.
In comparison, DSRL and RL Token achieve
77.78\% and 82.22\%, respectively.
Both baselines keep the base generative policy
frozen during online RL.
The performance gaps are particularly pronounced
on scissors manipulation and box opening,
while RL Token reaches 29/30 on screw insertion.

Fig.~\ref{fig:real_robot_tasks} shows
task-dependent online learning dynamics.
AE L14 recovers from an early decline on
scissors manipulation and follows similar
trends to Full FT on screw insertion.
On box opening, Full FT converges earlier,
while AE L14 eventually catches up.
Intervention rates decrease for both methods,
approaching zero on scissors manipulation
and screw insertion.
DSRL follows its original autonomous steering
setting without human intervention, while
RL Token shows persistent interventions
on box opening, potentially reflecting a
less reliable initial actor than the SFT policy.
These curves include human-assisted episodes,
whereas Tab.~\ref{tab:real_world_success}
reports independent final-policy evaluations.
Overall, single-layer adaptation matches
Full FT while outperforming frozen-policy
baselines across these tasks.
\section{Conclusion}
\label{sec:conclusion}

In this work, we investigate how much of a
pretrained VLA must remain trainable for
effective online reinforcement learning.
We conduct a systematic layer-wise study,
updating one original transformer block at
a time while keeping the remaining VLA
parameters frozen.
Our results show that several action expert
layers at higher relative depths achieve
performance comparable to full fine-tuning
while updating substantially fewer VLA parameters.
Further analyses show that individual modules
within a single layer can support adaptation
and that localized updates are accompanied by
attention redistribution across frozen layers.
We observe similar trends across different
VLA architectures and validate our findings
on three real-world precision manipulation tasks,
where single-layer adaptation matches full
fine-tuning performance.
Together, these results suggest that a single
transformer layer can provide sufficient capacity
for online VLA adaptation within the studied
settings.

\noindent\textbf{Acknowledgments.}
This work was supported in part by NSF CAREER,
NSF Grant No.~1941722, the RAI Institute,
ONR Grants N00014-22-1-2293 and N00014-22-1-2621,
the National Natural Science Foundation of China
(Grant No.~62476011), and the Beijing Natural
Science Foundation (Grant No.~L252060).

{
    \small
    \bibliographystyle{ieeenat_fullname}
    \bibliography{main}
}
\fi\clearpage\clearpage
\setcounter{page}{1}
\maketitlesupplementary
\renewcommand{\thepage}{S\arabic{page}}
\renewcommand{\thesection}{\Alph{section}}
\renewcommand{\theHsection}{appendix.\Alph{section}}
\setcounter{section}{0}
\renewcommand{\thetable}{S\arabic{table}}
\renewcommand{\theHtable}{S\arabic{table}}
\setcounter{table}{0}

\section{Implementation Details}

\begin{figure*}[t]
\centering
\includegraphics[width=\textwidth]{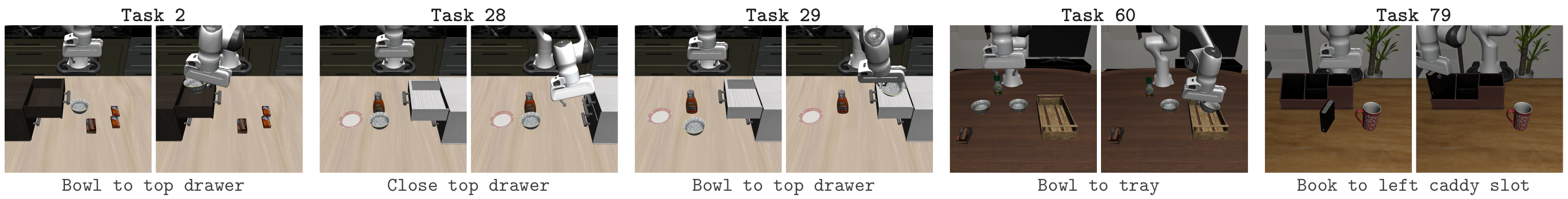}
\caption{Representative initial and goal scenes for the five LIBERO tasks.}
\label{fig:libero_five_task_scenes}
\end{figure*}

\label{app:online_implementation}

\begin{figure}[t] \centering \includegraphics[width=\columnwidth]{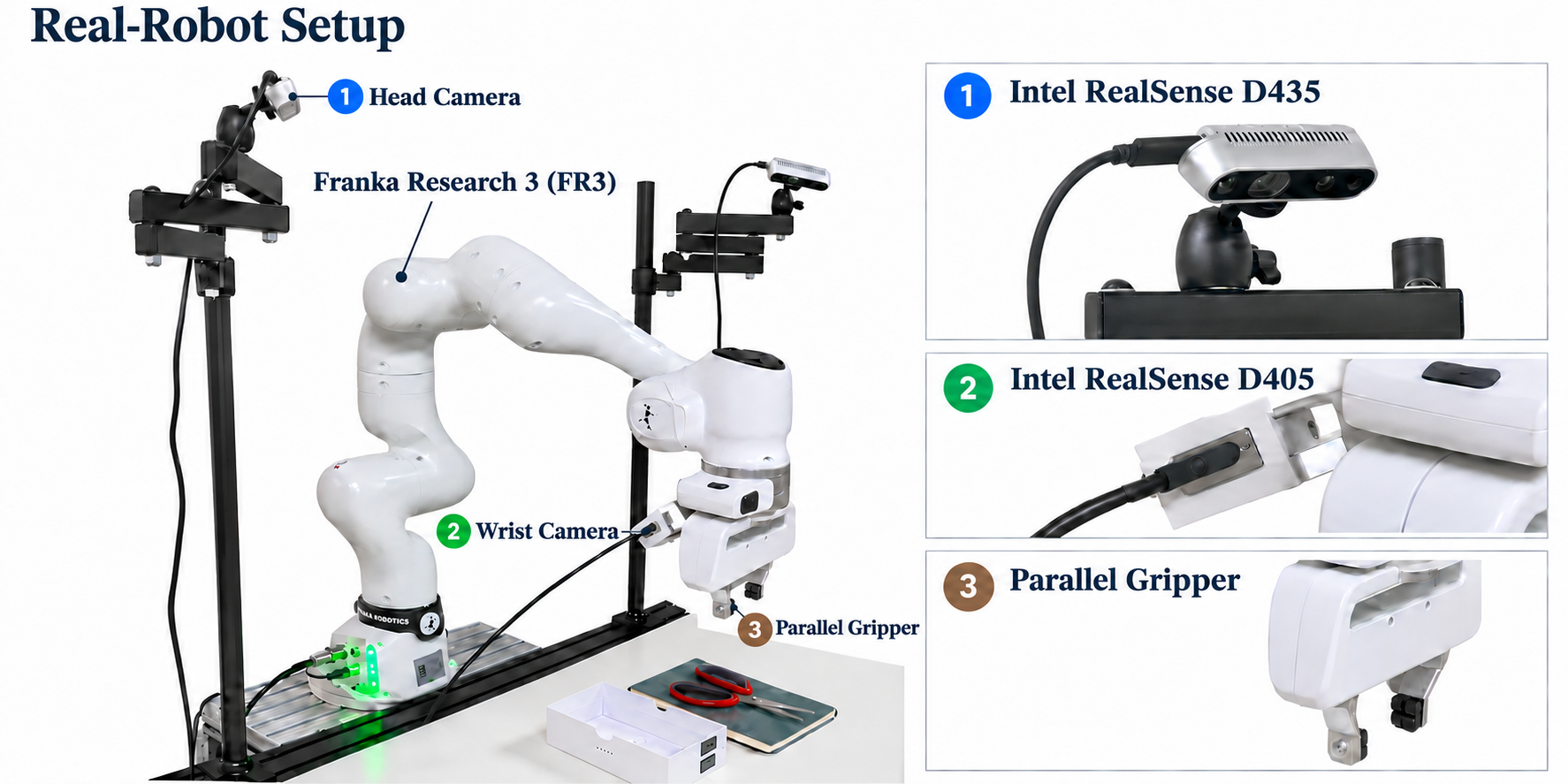} \caption{\textbf{Real-Robot Setup.} Franka Research 3 platform with head and wrist cameras and a parallel gripper.} \label{fig:real_robot_setup} \end{figure} \subsection{Optimization Hyperparameters}
\label{app:optimization_hyperparameters}

In simulation, actions are represented as delta end-effector (EEF) poses and gripper commands. Online RL uses only self-collected data, without task demonstrations. Following the original $\pi_{0.5}$ LIBERO configuration, we use a prediction horizon of 10 steps. Across backbones, we use the same replay buffer capacity, batch size, UTD ratio, and update schedule, while the execution horizon is 10 steps for $\pi_{0.5}$ and MolmoAct2 and 16 steps for GR00T-N1.6. Tab.~\ref{tab:sim_optimization_hyperparameters} summarizes the $\pi_{0.5}$ training hyperparameters. 
\begin{table}[t]
\centering
\caption{\textbf{Online adaptation settings for the
$\pi_{0.5}$ LIBERO experiments.}}
\label{tab:sim_optimization_hyperparameters}
\setlength{\tabcolsep}{5pt}
\resizebox{\columnwidth}{!}{%
\begin{tabular}{lc}
\hline
Setting & Value \\
\hline
Replay buffer capacity & 80,000 transitions \\ Training batch size & 32 \\

Chunk size & 10 steps \\
Execution horizon & 10 steps \\ Edit scale & 0.20 \\
Q-ensemble size / subsample & 10 / 2 \\
Critic / edit hidden layers & (256, 256, 256) \\
VLA / edited candidates per decision & 8 / 8 \\Updates per episode & 3 \\Optimizer & Adam \\ Critic / edit learning rates & $10^{-4}$ / $10^{-4}$ \\Discount & $0.99$ \\UTD ratio & 10 \\\hline
\end{tabular}}
\end{table}

\subsection{Cross-Backbone Experimental Details}
\label{app:cross_model_details}

This section provides architectural and
implementation details for the cross-backbone
experiments in Sec.~3.4 of the main paper.
All layer indices are zero-based. 
VLM depths refer to the language transformer
stack, excluding the vision encoder and
vision-language connector.

\noindent\textbf{$\pi_{0.5}$.} The Gemma and action expert streams have hidden widths of 2,048 and 1,024, respectively, and each contains 18 transformer blocks. The streams have separate parameters and interact through masked attention. Action tokens attend to visual-language context and other action tokens, while context tokens cannot attend to action tokens. The action expert uses timestep-conditioned adaptive RMS normalization and gated residual connections. \par \noindent\textbf{MolmoAct2.}~\cite{fang2026molmoact2actionreasoningmodels} Its VLM language stack and action expert each contain 36 blocks, with hidden sizes of 2,560 and 768, respectively. The main-paper comparison fine-tunes action expert block L27 (zero-based), including action self-attention, cross-attention to the corresponding VLM layer, the FFN, and timestep-conditioned adaptive RMS normalization and gates.  \par \noindent\textbf{GR00T-N1.6.}~\cite{nvidia2025gr00tn16} Its 32-block action DiT has hidden size 1,536 (32 attention heads of dimension 48), while the VLM context has hidden size 2,048. Its action blocks use timestep-conditioned adaptive LayerNorm. The implementation uses an Eagle-Block2A-2B-v2 backbone and shares VLM features across action layers. The main-paper comparison fine-tunes action block L26 (zero-based), including cross-attention to the VLM context, the FFN, and normalization and timestep-conditioning parameters. Odd-indexed blocks instead use self-attention over state and action representations.  \par \noindent\textbf{Trainable Parameters.} A single selected block contains approximately 23.60M parameters for $\pi_{0.5}$, 15.94M for MolmoAct2, and 34.62M for GR00T-N1.6 L26. Relative to the respective Full FT trainable VLA parameter counts of approximately 2,406.67M, 4,210M, and 1,890M, these correspond to 0.98\%, 0.38\%, and 1.83\%. These counts exclude auxiliary RL components.

\section{Additional Simulation Results}\label{app:additional_simulation_results}
%
%

\begin{figure*}[t] \centering \includegraphics[width=\textwidth]{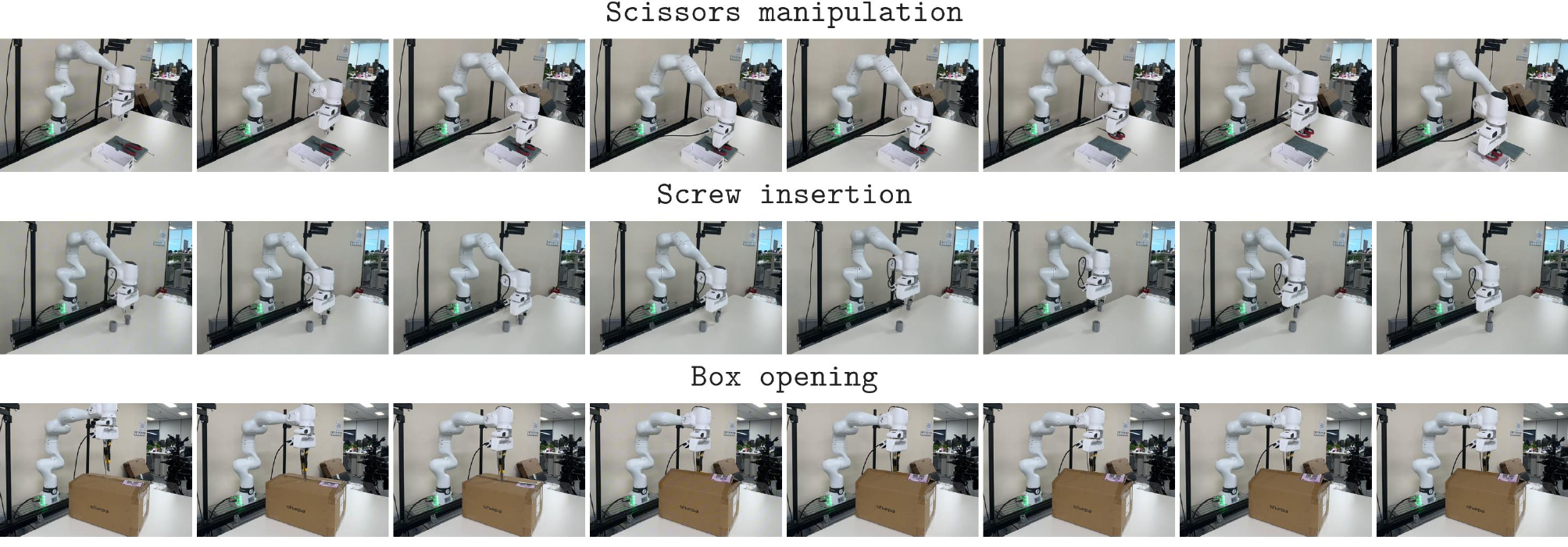} \caption{\textbf{Real-robot evaluation sequences.} Each row shows eight frames from one evaluation episode, ordered from left to right.} \label{fig:real_robot_evaluation_sequences} \end{figure*}

\subsection{Complete Layer-Wise Learning Curves}
\label{app:all_layer_curves} \providecommand{\Best}[1]{\textbf{#1}} \providecommand{\Second}[1]{\underline{#1}} \providecommand{\Third}[1]{\begingroup\setlength{\fboxsep}{0.3pt}\fbox{#1}\endgroup}

\begin{figure*}[t] \centering \includegraphics[ width=\textwidth, height=0.92\textheight, keepaspectratio ]{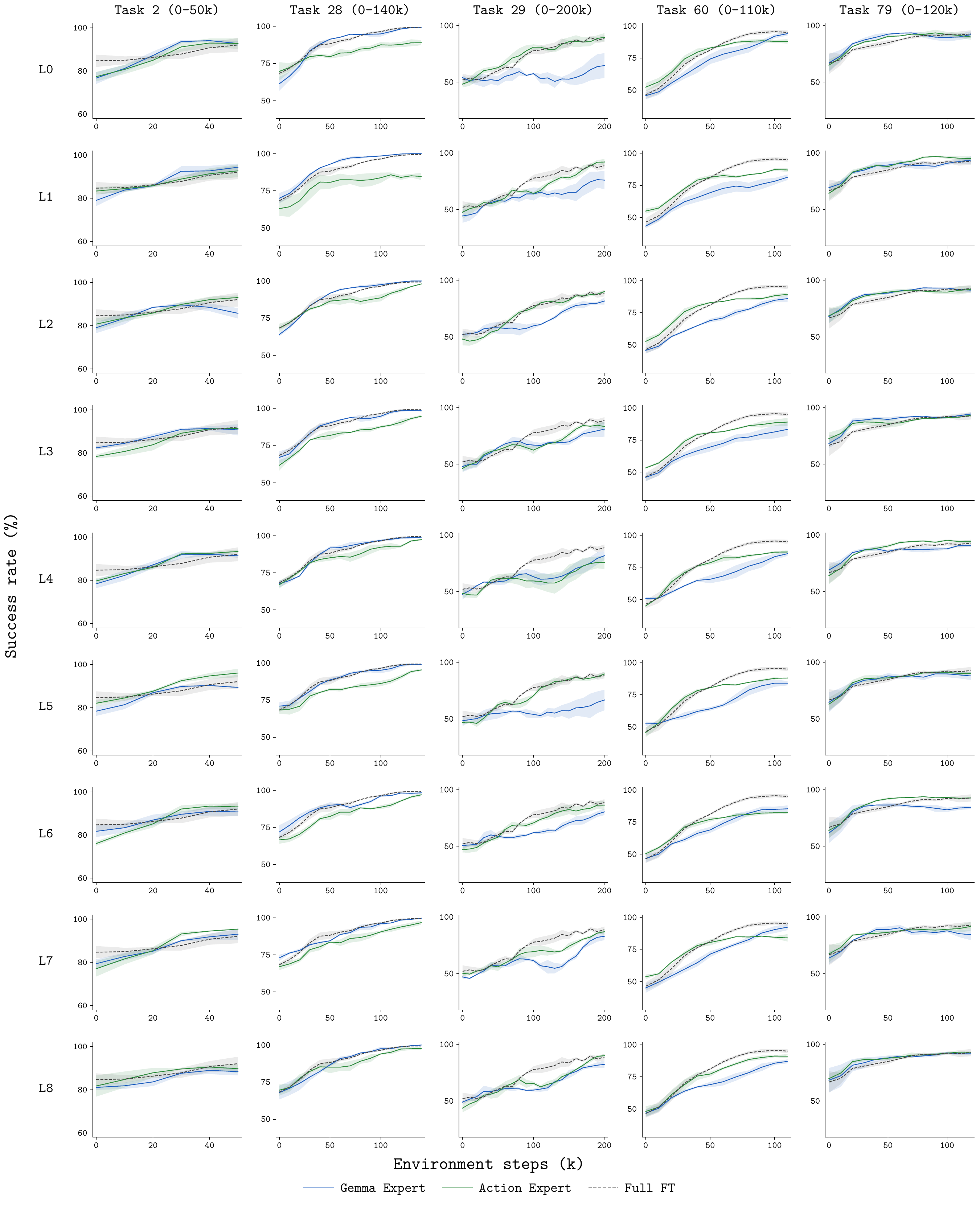} \caption{
    \textbf{Learning curves for single-layer adaptation at layers L0--L8 across five LIBERO tasks.}
    Curves show mean performance across three seeds with centered three-point smoothing,
    and shaded regions indicate pointwise standard errors across three seeds.
}
\label{fig:layer_curves_l0_l8} \end{figure*}

\begin{figure*}[t] \centering \includegraphics[ width=\textwidth, height=0.90\textheight, keepaspectratio ]{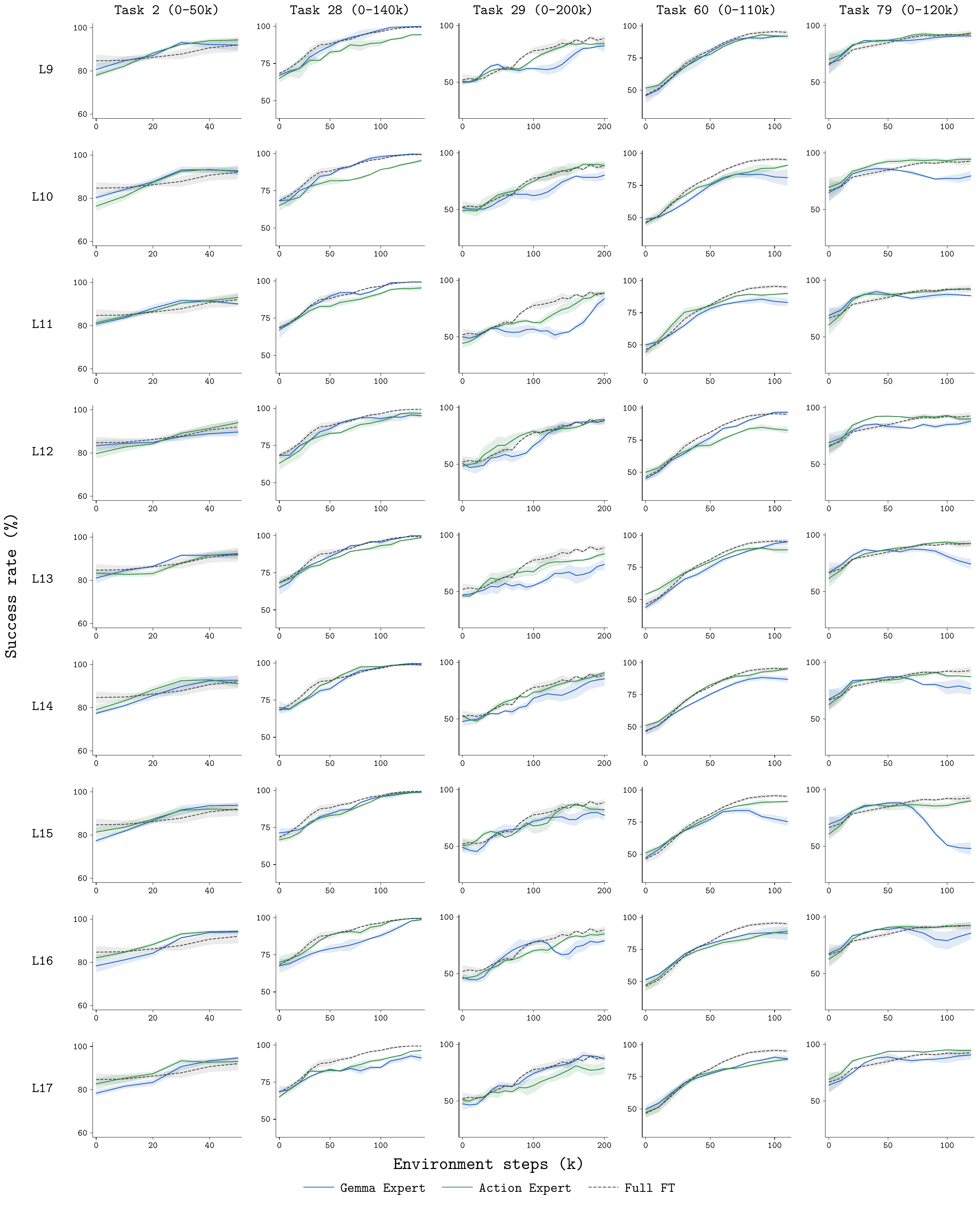} \caption{
    \textbf{Learning curves for single-layer adaptation at layers L9--L17 across five LIBERO tasks.}
    Curves show mean performance across three seeds with centered three-point smoothing,
    and shaded regions indicate pointwise standard errors across three seeds.
} \label{fig:layer_curves_l9_l17} \end{figure*}

Figs.~\ref{fig:layer_curves_l0_l8}
and~\ref{fig:layer_curves_l9_l17} present the
learning curves for single-layer adaptation
across layers L0--L17 of the Gemma and action
experts on five LIBERO tasks. Each curve shows
the evaluation success rate as a function of
environment steps, averaged over three seeds.
We apply a centered three-point moving average
to each seed before computing the mean and pointwise
standard error across three seeds. Shaded regions
indicate one standard error, while dashed lines denote
Full FT, which updates all transformer layers of both experts. Notably, action expert layers L13--L16 exhibit
similar learning dynamics and comparable final
success rates, suggesting that effective adaptation
is not limited to a single late layer.
We therefore select L14 as a representative
layer from this range.

%
\section{Real-Robot Experimental Details}\label{app:real_robot_details}
\subsection{Real-Robot Implementation Details}
\label{app:real_robot_implementation}   \noindent\textbf{Real-robot settings.} Fig.~\ref{fig:real_robot_setup} shows the robot platform and camera setup. We use 20 demonstrations for \textit{scissors manipulation} and 10 each for \textit{screw insertion} and \textit{box opening}. Baselines share the same task-specific demonstrations. Real-robot actions use absolute end-effector (EEF) poses. Offline demonstrations are collected at 10 Hz. Full FT and AE L14 use human interventions during training.  \begin{table}[t] \centering \caption{\textbf{Real-robot training settings.} } \label{tab:robot_optimization_hyperparameters} \setlength{\tabcolsep}{5pt} \begin{tabular*}{\columnwidth}{@{\extracolsep{\fill}}lc} \hline Setting & Value \\ \hline Demonstrations (scissors / screw / box) & 20 / 10 / 10 \\ Training batch size & 64 \\ Chunk size & 16 steps \\ Execution horizon & 10 steps \\ Q-ensemble size / subsample & 10 / 2 \\ Critic / edit hidden layers & (256, 256, 256) \\ VLA / edited candidates per decision & 8 / 8 \\ Updates per episode & 3 \\ Optimizer & Adam \\ Critic / edit learning rates & $10^{-4}$ / $10^{-4}$ \\ Discount & $0.99$ \\ UTD setting & 20 \\ \hline \end{tabular*} \end{table}

\subsection{Initial States and Reset Protocol} \label{app:real_robot_initial_states} The robot returns to the same task-specific reference pose between episodes. Object configurations vary across episodes, introducing variation in the observations and the alignment required for manipulation. \begin{itemize} \item \textit{Scissors manipulation.} Each episode starts with open scissors outside the target box. The scissors' orientation varies, changing the approach direction and gripper alignment to close and grasp them before~placement. \item \textit{Screw insertion.} The screw starts on the table, and the initial poses of the screw and the nut vary across episodes. These changes affect the grasp configuration and the alignment required for~insertion. \item \textit{Box opening.} The box orientation varies across episodes. The utility knife is held by the gripper at the beginning of each episode, so the policy begins with tool use rather than tool acquisition. Changes in box orientation alter the taped seam's direction relative to the blade and require adjustments to the cutting~motion. \end{itemize}

\noindent{
\textbf{Success criteria and evaluation.}
A trial is successful only if the task is completed
in full. For \textit{scissors manipulation}, success
requires the robot to close the scissors, grasp them
securely, and place them in the box. For \textit{screw
insertion}, success requires the screw to fall into
the nut and remain in its opening after the gripper
releases it. For \textit{box opening}, success requires
the blade to follow the taped seam and cut the tape
open. An operator judges each trial's success. 

For methods using human intervention during online
training, an operator judges when an episode is close
to failure and intervenes to provide corrective actions.
These corrections include aligning the screw with the
nut, aligning the gripper with the two finger holes of
the scissors, and adjusting the knife and box to align
the blade with the taped seam. Executed corrections
are stored in the replay buffer. Control automatically
returns to the policy when the human intervention ends.
Final evaluation uses 30 trials per task without human intervention.
}

\subsection{Real-Robot Baselines}
\label{app:real_robot_baselines}

\noindent\textbf{DSRL.}~\cite{wagenmaker2025steeringdiffusionpolicylatent}
We use a frozen, task-specific $\pi_{0.5}$ VLA
and train an RL actor to predict latent
noise, which the VLA transforms into an action chunk
conditioned on observations.
For \textit{scissors manipulation}, visual observations are
encoded into 128-dimensional features and concatenated
with proprioceptive states padded to 32 dimensions.
The actor predicts a 32-dimensional latent noise
vector, which is
broadcast across the VLA prediction horizon.

We employ an action critic and a noise critic, both
conditioned on the same visual and proprioceptive
features as the actor. The action critic evaluates
10-step action chunks with 7-dimensional actions,
flattened into 70-dimensional vectors. The noise
critic additionally takes the latent noise vector
as input and is trained using an MSE loss to match
the minimum of the twin action-critic Q-values for
the corresponding VLA-generated action chunk,
with gradients stopped through the regression target.
Both critics employ twin Q-networks. The actor and
all Q-networks use MLPs with hidden dimensions
$(256, 256, 256)$ and are optimized using Adam
with a learning rate of $3\times10^{-4}$. Following the autonomous online execution
setting in EXPO-FT~\cite{dong2026expoftsampleefficientreinforcementlearning},
we do not use human interventions during online training.

\noindent\textbf{RL Token.}~\cite{xu2026rltokenbootstrappingonline}
We adapt a third-party implementation of RL Token
to our real-robot setup. The readout projects
2048-dimensional image-token features from the
final VLA transformer layer to 512 dimensions
and aggregates them into an RL token using a
learnable query and cross-attention. The readout
is trained to reconstruct the original features
using a non-autoregressive decoder with a masked
MSE loss. We train
the readout for 2,000 gradient updates using
Adam with a learning rate of $10^{-4}$. During online RL, the VLA and readout remain frozen.
A Gaussian actor conditions on the RL token,
7-dimensional proprioceptive state, and a
VLA-predicted reference action chunk of shape
$(10, 7)$. The actor and twin Q-networks use
MLPs with hidden dimensions $(256, 256)$.
The actor objective combines Q-value maximization,
action regularization, and adjacent-step
pose-difference matching. We optimize the actor
and critic using Adam with a learning rate of
$10^{-4}$, a batch size of 64, and a discount
factor of $0.99$. \clearpage   \ifarxiv\else
\clearpage
\bibliographystyle{ieeenat_fullname}
\bibliography{main}

\fi\end{document}